\documentclass{article}
\usepackage{times}
\usepackage{algorithm,algorithmic,a4wide,amssymb}
\begin{document}
\def\GM{G\"{o}del Machine }
\def\gm{G\"{o}del machine }
\def\GMn{G\"{o}del Machine}
\def\gmn{G\"{o}del machine}
\def\hs{Hsearch }
\date{}

\title{{\sc PowerPlay:} Training an Increasingly \\ General Problem Solver by Continually Searching \\ for the Simplest Still Unsolvable Problem}

\date{22 December 2011 (arXiv:1112.5309v1), revised 4 November 2012}
\author{J\"{u}rgen Schmidhuber \\
The Swiss AI Lab IDSIA, Galleria 2, 6928 Manno-Lugano \\
University of Lugano \& SUPSI, Switzerland}
\maketitle

\begin{abstract}
Most of computer science focuses on automatically solving given computational problems.
I focus on automatically {\em inventing} or {\em discovering} problems in a way inspired by the playful behavior of 
animals and humans,
to train a more and more general problem solver from scratch in an unsupervised fashion.
Consider the infinite set of all computable descriptions of tasks with possibly computable solutions.
Given a general problem solving architecture,
at any given time, the novel algorithmic framework {\sc PowerPlay}  \cite{Schmidhuber:11powerplay} searches 
the space of possible {\em pairs} of new tasks
and modifications of the current problem solver, 
until it finds a more powerful problem solver that provably
solves all previously learned tasks plus the new one, 
while the unmodified predecessor does not. 
Newly invented tasks may require to achieve a {\em wow-effect} by
making previously learned skills more efficient such that
they require less time and space.
New skills may (partially) re-use previously learned skills.
The greedy search of typical
 {\sc PowerPlay} variants  uses time-optimal program search to order
candidate pairs of tasks and solver modifications by their conditional computational (time \& space) complexity,
given the stored experience so far. 
The new task and its corresponding task-solving skill are those first found and validated.
This biases the search towards
pairs that can be described compactly and validated quickly.
The computational costs of validating new tasks need not grow with task repertoire size.
Standard  problem solver architectures of personal computers  or 
neural networks tend to generalize by solving numerous tasks outside the self-invented training set;
{\sc PowerPlay}'s ongoing search  for novelty keeps breaking the generalization abilities
of its present solver.
This is related  
 to G\"{o}del's sequence of increasingly powerful formal theories
based on adding formerly
unprovable statements  to the axioms without affecting previously provable theorems.
The continually increasing repertoire of problem
solving procedures can be exploited by a parallel search for solutions to additional externally posed tasks.
{\sc PowerPlay} may be viewed as a greedy but practical implementation of basic principles of creativity
\cite{Schmidhuber:06cs,Schmidhuber:10ieeetamd}.
A first experimental analysis can be found in separate papers \cite{rupesh2012icdl,powerplay2012experiments}.

\end{abstract}

\newpage
\tableofcontents

\newpage
\section{Introduction}
\label{intro}

Given a realistic piece of computational hardware with specific resource limitations, 
how can one devise software for it that will solve all, or at least many,
of the {\em a priori} unknown tasks that are in principle easily solvable on this architecture?
In other words, how to build a {\em practical} general problem solver, given the computational restrictions?
It does not need to be {\em universal} and  {\em asymptotically} optimal 
\cite{Levin:73,Hutter:01fast+,Schmidhuber:04oops,Schmidhuber:09gm}  
like the recent (not necessarily practically feasible) general problem solvers discussed
in Section \ref{optimal};  instead it should take into account all
constant architecture-specific slowdowns  ignored in the asymptotic 
optimality notation of theoretical computer science,
and be generally useful for real-world applications.

Let us draw inspiration from biology.
How do initially helpless human babies become rather general problem solvers over time?
Apparently by playing. For example, even in the absence of external reward or hunger they are curious about what happens if they move their eyes or fingers in particular ways, creating little experiments which lead to initially novel and surprising but eventually predictable sensory inputs, while also  learning motor skills to reproduce these outcomes. (See \cite{Schmidhuber:90sab,Schmidhuber:91singaporecur,Schmidhuber:99cec,Schmidhuber:06cs,Schmidhuber:10ieeetamd,sunyi2011agi} and Section \ref{creativity} for previous artificial systems of this type.) Infants continually seem to invent new tasks that become boring as soon as their solutions become  known. Easy-to-learn new tasks are preferred over unsolvable or hard-to-learn tasks. Eventually the numerous skills acquired in this creative, self-supervised way may get re-used to facilitate the search for solutions to  external problems, such as finding food when hungry.
While kids keep inventing new problems for themselves, they move through remarkable developmental stages \cite{Piaget:55,Berlyne:54,Harlow:50}.

Here I introduce a novel unsupervised algorithmic framework for 
training a computational problem solver from scratch, continually searching
for the simplest (fastest to find) combination of 
 task and corresponding task-solving skill to add to its growing repertoire, 
without forgetting any previous skills (Section \ref{framework}),
 or at least without decreasing average performance on previously solved tasks (Section  \ref{costsection}).
New skills may (partially) re-use previously learned skills.
Every new task added to the repertoire is essentially defined by the time required 
to invent it, to solve it, and to demonstrate that no previously learned skills got lost.
The search 
takes into account that typical problem solvers may learn to solve tasks outside the growing self-made training set 
due to generalization properties of their architectures. 
The framework is called {\sc PowerPlay}  because it continually \cite{Ring:94}
aims at boosting computational prowess and problem solving capacity,
reminiscent of humans or human societies 
trying to boost their general power/capabilities/knowledge/skills  in playful ways, even in the absence of
externally defined goals, although 
the skills learned by this type of pure curiosity may later help to solve 
externally posed tasks.

Unlike our first implementations of curious/creative/playful agents
from the 1990s  \cite{Schmidhuber:91singaporecur,Storck:95,Schmidhuber:99cec} (Section \ref{creativity}; compare
  \cite{Barto:12intrinsic,Dayan:12intrinsic,Oudeyer:12intrinsic,Nemzhov:12intrinsic}),
{\sc PowerPlay} provably (by design) does not have any problems with online learning---it cannot forget previously learned 
skills, automatically segmenting its life into a sequence of clearly identified tasks with explicitly recorded solutions.
Unlike the task search of theoretically optimal creative agents  \cite{Schmidhuber:06cs,Schmidhuber:10ieeetamd} 
(Section \ref{creativity}), {\sc PowerPlay}'s
task search is greedy, but at least practically feasible.

Some claim that scientists often invent appropriate problems for their methods,
rather than inventing methods to solve given problems. The present paper
formalizes this in a way that may be more convenient to implement
than those of previous work \cite{Schmidhuber:91singaporecur,Schmidhuber:99cec,Schmidhuber:06cs,Schmidhuber:10ieeetamd}, 
and describes a simple practical framework for building
creative artificial scientists or explorers that by design continually 
come up with the fastest to find, initially novel, but eventually solvable problems.

\subsection{Basic Ideas}
\label{idea}

In traditional computer science, given some formally defined task, 
a search algorithm is used to search a space of solution 
candidates until a solution to the task is found and verified.
If the task is hard the search may take long.

To automatically construct an increasingly general problem solver,
let us expand the traditional search space in an unusual way, 
such that it includes all possible {\em pairs} of computable tasks with possibly computable solutions, and problem solvers.
Given an old problem solver that can already solve a finite known set of previously learned tasks,
a search algorithm is used to find a new pair that provably has the following properties:
{\bf (1)} The new task cannot be solved by the old problem solver.
{\bf (2)} The new task can be solved by the new problem solver (some modification of the old one).
{\bf (3)}  The new solver can still solve the known set of previously learned tasks.

Once such a pair is found, the cycle repeats itself. This will result in a continually 
growing set of known tasks solvable by an increasingly more powerful problem solver.
 Solutions to new tasks may (partially) re-use solutions to previously learned tasks.

Smart search (e.g., Section \ref{oops}, Algorithm  \ref{dooops})
orders candidate pairs of the type {\em (task, solver)}
 by computational complexity, using concepts of 
optimal universal search \cite{Levin:73,Schmidhuber:04oops},
 with a bias towards pairs that can be described by few additional
bits of information (given the experience so far)
and that can be validated quickly.

At first glance it might seem harder to search for pairs of tasks and solvers instead of solvers only,
due to the apparently larger search space. However, the additional freedom of {\em inventing} the
tasks to be solved may actually greatly reduce the time 
intervals between problem solver advances, because the system may often have the option
of inventing a rather simple task with an easy-to-find solution.

A new task may be about simplifying the old solver such that it can still solve all tasks learned so far,
but with less computational resources such as time and storage space (e.g., 
Section \ref{taskinvention} and Algorithm \ref{cost}).

Since the new pair {\em (task, solver)} is the first one found 
and validated, the search automatically trades off the time-varying efforts required 
to either invent completely new, previously unsolvable problems, 
or compressing/speeding up previous solutions. 
Sometimes it is easier to refine or simplify known skills, sometimes to invent new skills.

On typical problem solver architectures of personal computers (PCs) or neural networks (NNs), 
while a limited known number of previously learned tasks has become solvable,
so too has a large number of unknown, never-tested tasks (in the field of Machine Learning, this is known as {\em generalization}). 
 {\sc PowerPlay}'s ongoing search is continually testing (and always trying to go beyond) 
the generalization abilities of the most recent solver instance;
some of its search time has to be spent on demonstrating that  self-invented new tasks
are not  already solvable.

Often, however, much more  time will have to be spent on making sure that a newly modified solver
did not forget any of the possibly many previously learned skills.
Problem solver modularization (Section \ref{correctness}, especially \ref{components}) 
may greatly reduce this time though, making 
 {\sc PowerPlay} prefer pairs whose validation does not require the re-testing of
too many previously learned skills, thus decomposing at least part of the search space into 
somewhat independent regions, realizing {\em divide and conquer} strategies as by-products of 
its built-in drive to invent and validate novel tasks/skills as quickly as possible.


A biologically inspired hope is that as the problem solver is becoming more and more general,
it will find it easier and easier to solve externally posed tasks (Section \ref{external}), just like
growing infants often seem to  re-use their playfully acquired skills to solve 
teacher-given problems.

\subsection{Outline of Remainder}

Section  \ref{framework} will introduce basic notation and Variant 1 of the algorithmic framework
{\sc PowerPlay}, which invokes the essential 
procedures {\sc Task Invention},  {\sc Solver Modification}, and {\sc Correctness Demonstration}.
Section  \ref{more} will discuss details of these procedures.

More detailed instantiations of {\sc PowerPlay} will be described in 
Section \ref{evolution} (an evolutionary method, Alg. \ref{doevo}) and 
Section \ref{oops} (an asymptotically optimal program search method, Alg.  \ref{dooops}).

As mentioned above, the skills acquired to solve
self-generated tasks may later greatly facilitate solutions to externally posed tasks, 
just like the numerous motor skills learned by babies during curious exploration of its world often
can be re-used later to maximize external reward.
Sections \ref{external} and \ref{cost} will discuss variants of the framework (e.g., Algorithm \ref{cost})  in which some
of the tasks can be defined externally. 

Section  \ref{costsection}  will also describe
a natural variant of the framework that explicitly penalizes solution costs (including time and space complexity),
and  allows for forgetting aspects of previous solutions, provided the average performance on
previously solved tasks does not decrease.

Section \ref{experiments} will point to illustrative experiments
in separate papers \cite{rupesh2012icdl,powerplay2012experiments}.
Section \ref{previous} will discuss relations to previous work.

\section{Notation \& Algorithmic Framework {\sc PowerPlay} (Variant I)}
\label{framework}

$B^*$ denotes the set of finite
sequences or bitstrings over the binary alphabet $B=\{0,1\}$, 
 $\lambda$ the empty string,
 $x,y,z,p,q,r,u$  strings in $B^*$,
 $\mathbb{N}$ the natural numbers,
 $\mathbb{R}$ the real numbers,
 $\epsilon \in \mathbb{R}$ a positive constant,
  $m, n,n_0,k,i,j,k,l$ non-negative integers,
$L(x)$ the number of bits in $x$ (where  $L(\lambda) = 0$),
$f,g$
functions mapping integers to integers.  We write $f(n)=O(g(n))$ if
there exist positive $c,n_0$ such that $f(n) \leq cg(n)$ for
all $n > n_0$.  

The computational architecture of the problem solver may be a deterministic universal computer, 
or a more limited device such as a finite state automaton or a feedforward neural network (NN) \cite{bishop:2006}.
All such problem solvers can be uniquely encoded \cite{Goedel:31} or implemented on universal computers  such as 
universal Turing Machines (TM) \cite{Turing:36}. Therefore,
without loss of generality, the remainder of this paper 
assumes a fixed universal reference computer
whose input programs and outputs are elements of $B^*$.  
A user-defined subset  $\cal S $ $\subset B^*$ defines the set of possible problem solvers. 
For example, if the problem solver's architecture is itself a binary universal TM or a standard 
computer,  then $\cal S$ represents its set of possible programs, or a limited subset thereof---compare Sections \ref{policychange} and \ref{oops}.
If it is a feedforward NN, then $\cal S$ could be a highly restricted subset of programs  
encoding the NN's possible topologies and weights (floating point numbers)---compare Section \ref{experiments}
and the original SLIM NN paper \cite{Schmidhuber:12slimnn}.

In what follows, for convenience
I will often identify bitstrings  in $B^*$ with things  they encode, such
as integers, real-valued vectors, weight matrices, or programs---the context will always
make clear what is meant.

The problem solver's initial program is called $s_0$.
There is  a set of possible task descriptions $\cal T$ $ \subset B^*$.
$\cal T$ may be the infinite set of {\em all} possible computable descriptions of tasks with possibly computable solutions,
or just a small subset thereof.
For example, a simple task may require the solver to answer a particular input pattern
with a particular output pattern (more formal details on
pattern recognition tasks are given in Section \ref{patternrecognition}).
Or it may require the solver to steer a robot towards a goal through a sequence of actions
(more formal details on sequential decision making tasks in unknown environments 
are given in Section \ref{decisionmaking}).
There is a particular sequence of task descriptions $T_1, T_2, \ldots$, 
where each unique $T_i  \in \cal T$ $(i=1, 2, \ldots)$
is chosen or ``invented'' by a search method described below
such that the solutions of  $T_1, T_2, \ldots, T_i$ 
can be computed by $s_i$, the $i$-th instance of the program, 
but not by $s_{i-1}$ $(i=1, 2, \ldots)$.
Each $T_i$ consists of a unique problem identifier
that can be read by  $s_i$ through some built-in input processing mechanism 
(e.g.,  input neurons of an NN  \cite{Schmidhuber:12slimnn}), 
and a unique description of a deterministic procedure for determining 
whether the problem has been solved.
Denote $T_{\leq i}= \{T_1, \ldots, T_i\}$;
$T_{< i}= \{T_1, \ldots, T_{i-1}\}$.

A valid task $T_i(i>1)$ may 
require solving at least one previously solved task $T_k (k<i)$ more efficiently, by using less
resources such as storage space, computation time, energy, etc.,
thus achieving a {\em wow-effect}.
See Section \ref{taskinvention}.

Tasks and problem solver modifications  are computed and validated by elements of 
another appropriate set of programs $\cal P$ $\subset B^*$. Programs $p \in \cal P$
may contain instructions for reading and executing (parts of) the code of the present problem solver and reading (parts of) a recorded
history $ Trace \in B^*$ of previous events that led to the present solver. 
The algorithmic framework (Alg. \ref{powergrab})
incrementally trains the problem solver by finding $p \in \cal P$ that increase the set of solvable tasks.

\begin{algorithm}{\bf Alg. \ref{powergrab}: Algorithmic Framework {\sc PowerPlay} (Variant I)} 
\label{powergrab}
\begin{algorithmic}
\STATE Initialize $s_0$ in some way.
\FOR {$i := 1, 2, \ldots$} 
\REPEAT 
\STATE Let a search algorithm (examples in Section \ref{implement}) create a new candidate program $p  \in \cal P$. Give $p$ limited time to do (not necessarily in this order):
 \STATE    {\bf *} {\sc Task Invention}: Let $p$ compute a task $T \in \cal T$. See Section \ref{taskinvention}.
\STATE  {\bf *}  {\sc Solver Modification}: Let $p$ compute a value of the variable $q \in \cal S$ $ \subset B^*$  (a candidate for  $s_i$) by computing a modification of $s_{i-1}$. See Section \ref{policychange}.
\STATE   {\bf *}  {\sc Correctness Demonstration}: Let $p$ try to show that $T$ cannot be solved by $s_{i-1}$, but that $T$ and all $T_k (k<i)$ can be solved by  $q$.  See Section \ref{correctness}.
\UNTIL {{\sc Correctness Demonstration} was successful}
\STATE Set $p_i := p; T_i := T; s_i := q$; update $Trace$.
\ENDFOR
\end{algorithmic}
\end{algorithm}

\section{{\sc Task Invention},  {\sc Solver Modification},  {\sc Correctness Demo}}
\label{more}

A program tested by Alg. \ref{powergrab} has to allocate its runtime  to solve three main jobs, namely,
{\sc Task Invention},  {\sc Solver Modification},  {\sc Correctness Demonstration}.
Now examples of each will be listed.

\subsection{Implementing {\sc Task Invention}}
\label{taskinvention}

Part of the job of $p_i \in \cal P$ is to compute $T_i \in \cal T$. 
This will consume some of the total computation time allocated to $p_i$.
Two examples will be given:
pattern recognition tasks are treated in Section \ref{patternrecognition};
sequential decision making tasks in Section \ref{decisionmaking}.

\subsubsection{Example: Pattern Recognition Tasks} 
\label{patternrecognition}

In the context of  learning to recognize or analyze patterns,
$T_i$ could be a 4-tuple $(I_i, O_i, t_i, n_i) \in \cal I \times \cal O \times 
\mathbb{N} \times \mathbb{N} $, 
where  $\cal I, O \subset B^*$, and
$T_i$ is solved if $s_i$ satisfies $L(s_i)< n_i$ and
needs at most $t_i$ discrete time steps  to read $I_i$ and compute $O_i$ and halt.
Here $I_i$ itself may be a pair $(I^1_i, I^2_i) \in B^* \times \cal B^* $, where 
$I^1_i$ is constrained to be the address of an image in a given database of patterns,
and $I^2_i$ is a  $p_i$-generated {\em ``query''}
that uniquely specifies {\em how} the image 
should be classified through target pattern $O_i$,
such that the same image can be analyzed in different ways
during different tasks.
For example, depending on the nature of the invented task sequence,
the problem solver could eventually learn that $O=1$  if $I^2=1001$ (suppressing task indices)
and the image addressed by $I^1$ contains at least one black pixel, 
or   if $I^2=0111$ and the image shows a cow.

Since the definition of task $T_i$ includes bounds $n_i,t_i$ on computational resources,
$T_i$ may be about solving at least one $T_k (k<i)$ more efficiently,
corresponding to a {\em wow-effect}.
This in turn may also yield more efficient solutions to other tasks $T_l (l<i,l \neq k)$.
In practical applications one may insist that such efficiency gains must exceed a certain
threshold $\epsilon>0$, to avoid task series causing sequences of very minor improvements.

Note that $n_i$ and $t_i$ may be unnecessary in 
special cases such as the problem solver being a fixed topology feedforward
NN \cite{bishop:2006} 
whose input and target patterns have constant size and whose
computational efforts per pattern need constant time and space resources.

Assuming  sufficiently powerful  $\cal S, \cal P$,
in the beginning,
trivial tasks such as simply copying $I^2_i$ onto $O_i$ may be interesting 
in the sense that {\sc PowerPlay} can still validate and accept them,
but they will become boring (inadmissible) 
as soon as they are solvable by solutions to previous tasks that generalize
to new tasks.

\subsubsection{Example: General Decision Making Tasks in Dynamic Environments} 
\label{decisionmaking}

In the more general context of general problem solving/sequential decision making/reinforcement learning/reward optimization \cite{Newell:63,Kaelbling:96,suba98} in unknown environments, 
there may be a set $\cal I$ $\subset B^*$ of possible task identification patterns and
a set $\cal J$ $ \subset B^*$ of programs that test properties of bitstrings.
$T_i$ could then encode a 4-tuple $(I_i, J_i, t_i, n_i)\in \cal I \times \cal J \times \mathbb{N} \times \mathbb{N}$ of finite bitstrings
with the following interpretation:
$s_i$ must satisfy $L(s_i)< n_i$ and may spend at most $t_i$ discrete time steps  on first reading $I_i$ 
and then interacting with an environment through a sequence of perceptions and actions,
to achieve some computable goal defined by $J_i$.

More precisely, while $T_i$ is being solved within $t_i$ time steps,
at any given time $1 \leq t \leq t_i$,
the internal state of the problem solver at time $t$ is denoted $u_i(t) \in B^*$; its initial default value is $u_i(0)$.
For example, $u_i(t)$ may encode the current contents of the internal tape of a TM, 
or of certain addresses in the dynamic storage area of a PC,
or the present activations of an LSTM recurrent NN \cite{Hochreiter:97lstm}.
At time $t$, $s_i$ can spend a constant number of elementary computational instructions to copy
the task dscription $T_i$ or 
the present environmental input $x_i(t) \in B^*$ and a reward signal $r_i(t) \in B^*$ (interpreted
as a real number)
into parts of $u_i(t)$, then update 
other parts of $u_i(t)$ (a function of 
$u_i(t-1)$) and compute
action $y_i(t) \in B^*$ encoded as a part of $u_i(t)$. $y_i(t)$ may affect the environment, and thus future inputs. 

If $\cal P$ allows for programs that can {\em dynamically acquire additional
physical computational resources} such as additional CPUs and storage,
then the above constant number of elementary computational instructions
should be replaced by a constant amount of real time, to be measured by a reliable physical clock.

The sequence of 4-tuples $(x_i(t),r_i(t),u_i(t),y_i(t))$ $(t=1,\ldots,t_i)$
gets  recorded by the so-called trace $Trace_i \in B^*$.
If at the end of the interaction a desirable computable property $J_i (Trace_i)$  (computed by applying
program $J_i$ to  $Trace_i$)  is satisfied, then by definition the task is solved.
The set  $\cal J$ of  possible $J_i$ may represent an infinite set of all computable tasks with solutions computable
by the given hardware.
For practical reasons, however,
the set  $\cal J$ of  possible $J_i$ may also be restricted to bit sequences encoding just a few possible goals.
For example,  $J_i$ may only encode goals of the form:
a robot arm steered by program or {\em``policy''} $s_i$ has reached a certain target (a desired final observation $x_i(t_i)$ 
recorded in $Trace_i$) 
without measurably bumping into an obstacle along the way, that is, there were no negative rewards, that is, 
$  r_i(\tau) \geq 0$ for $\tau=1 \ldots t_i$.

If the environment is deterministic, e.g., a digital physics simulation of a robot, 
then its current state can be encoded as part of $u(t)$,
and it is straight-forward for {\sc Correctness Demonstration}
to test whether some $s_i$ still can solve a previously solved task $T_j (j<i)$.
However, what 
if the environment  is only partially observable, like the real world, and non-stationary, 
changing in unknown ways?
Then {\sc Correctness Demonstration} must check whether $s_i$ still produces the same action sequence in response to
the input sequence recorded in $Trace_j$ (often this replay-based test will actually be computationally cheaper than
a test involving the environment). 
{\em Achieving the same goal in a changed environment  must be considered a different task,
even if the changes are just due to noise on the environmental inputs.}
(Sure, in the real world $s_j (j>i)$ might actually 
achieve $J_i$ faster than $s_i$, given the description of $T_i$, but   {\sc Correctness Demonstration} in
general cannot know whether this acceleration was due to plain luck---it must stick to reproducing $Trace_j$
to make sure it did not forget anything.)

See Section \ref{probabilistic}, however, for a less strict {\sc PowerPlay} variant whose {\sc Correctness Demonstration} directly interacts with the real world to collect sufficient problem-solving statistics through repeated trials, making certain
assumptions about the probabilistic nature of the environment, and the repeatability of  experiments.

\subsection{Implementing {\sc Solver Modification}}
\label{policychange}

Part of the job of $p_i \in \cal P$ is also to compute $s_i$, possibly profiting from having access to $s_{i-1}$, because only few changes of  $s_{i-1}$ may be necessary to come up with an $s_i$ that goes beyond $s_{i-1}$. For example, if  the problem solver is a standard PC,  then just a few bits of the previous software $s_{i-1}$ may need to be changed.

For practical reasons, the set  $\cal S$ of  possible $s_i$ may be greatly restricted to bit sequences 
encoding programs that obey the syntax of a standard programming language such as LISP or Java. In turn, the programming language describing  $\cal P$ should be greatly restricted such that any $p_i \in \cal P$ can only produce syntactically correct $s_i$.

If the problem solver is a feedforward NN with pre-wired, unmodifiable topology, then 
$\cal S$ will be restricted to those bit sequences encoding valid weight matrices, 
$s_i$ will encode its $i$-th weight matrix, and $\cal P$ will be restricted to those $p \in \cal P$ 
that can produce  some $s_i \in \cal S$. Depending on the user-defined programming language,
$p_i$ may invoke complex pre-wired subprograms (e.g., well-known
learning algorithms) as primitive instructions---compare separate experimental analysis \cite{rupesh2012icdl,powerplay2012experiments}.

In general, $p$ itself determines how much time to spend on {\sc Solver Modification}---enough time must be left
for {\sc Task Invention} and {\sc Correctness Demonstration}.

\subsection{Implementing {\sc Correctness Demonstration}}
\label{correctness}

Correctness demonstration may be the most  time-consuming obligation of $p_i$.
At first glance it may seem that as the sequence $T_1, T_2, \ldots $ is growing,
more and more time will be needed to show that $s_i$ but not $s_{i-1}$ can solve $T_1, T_2, \ldots, T_i$,
because one naive way of ensuring correctness is to re-test $s_i$ on all previously solved tasks.
Theoretically more efficient ways are considered next.

\subsubsection{Most General: Proof Search}
\label{proofsearch}

The most general way of demonstrating correctness is to
encode  (in read-only storage) an axiomatic system $\cal A$  that
formally describes computational properties of the problem solver and possible $s_i$, 
and to allow $p_i$ to search the space of possible proofs derivable from $\cal A$, using
a proof searcher subroutine 
that systematically generates proofs 
until it finds a  theorem stating that $s_i$ but not $s_{i-1}$ solves $T_1, T_2, \ldots, T_i$ (proof search may 
achieve this efficiently without explicitly re-testing $s_i$ on $T_1, T_2, \ldots, T_i$).
This could be done like in the \GM \cite{Schmidhuber:09gm} (Section \ref{optimal}), which 
uses an online extension of {\em Universal Search} 
\cite{Levin:73}
to systematically test {\em 
proof techniques}: proof-generating programs that
 may invoke special instructions
for generating axioms and applying inference rules to prolong an
initially empty {\em proof} $\in B^*$ by theorems, which are
either axioms or inferred from previous
theorems through rules such
as {\em modus ponens} combined with {\em unification}, e.g.,
\cite{Fitting:96}.  
$\cal P$ can be easily  limited to programs generating only syntactically correct proofs \cite{Schmidhuber:09gm}.
$\cal A$  has to subsume axioms describing 
how any instruction invoked
by some $s \in \cal S$ will change 
the state $u$ of the problem solver
from one step to the next (such that proof techniques can reason
about the effects of any $s_i$). Other axioms encode knowledge about arithmetics etc
(such that proof techniques can reason about spatial and temporal resources consumed by $s_i$).

In what follows,  {\sc Correctness Demonstrations} will be discussed
that are less general but sometimes more convenient to implement.

\subsubsection{Keeping Track Which Components of the Solver Affect Which Tasks}
\label{components}

Often it is possible to
partition $s \in \cal S$ into components, such as individual bits of the software of a PC,
or weights of a NN.
Here the $k$-th component of $s$ is denoted $s^k$.
For each $k$ $(k=1, 2, \ldots)$  a variable list $L^k=(T^k_1, T^k_2, \ldots)$ is introduced.
Its initial value before the start of {\sc PowerPlay} is $L^k_0$, an empty list.
Whenever $p_i$ found $s_i$ and $T_i$ at the end  of  {\sc Correctness Demonstration}, 
each $L^k$ is updated as follows: Its new value $L^k_i$ is obtained by appending to $L^k_{i-1}$
those $T_j \notin L^k_{i-1} (j=1,  \ldots, i)$ whose current (possibly revised) solutions now need  $s^k$ at 
least once during the solution-computing process,
and deleting those $T_j$ whose current solutions do not use  $s^k$ any more.

{\sc PowerPlay}'s {\sc Correctness Demonstration} thus has to test only tasks in the union of all $L^k_i$.
That is, if the most recent task does not require changes of many components of $s$, and 
if the changed bits do not affect many previous tasks, then  {\sc Correctness Demonstration}
may be very efficient.

Since every new task added to the repertoire is essentially defined by the time required 
to invent it, to solve it, and to show that no previous tasks became unsolvable in the process,
{\sc PowerPlay} is generally ``motivated'' to invent tasks whose validity check does not require too much
computational effort. That is, {\sc PowerPlay} will often find $p_i$ that generate $s_{i-1}$-modifications that don't
affect too many previous tasks, thus decomposing at least part of the spaces of tasks and their solutions  into more or
less independent regions, realizing {\em divide and conquer} strategies as by-products. Compare a recent experimental
analysis of this effect \cite{rupesh2012icdl,powerplay2012experiments}.

\subsubsection{Advantages of Prefix Code-Based Problem Solvers}
\label{prefix}

Let us restrict $\cal P$ such that tested $p \in \cal P$ cannot change any components of $s_{i-1}$ during {\sc Solver Modification},
but can create a new $s_i$ only by adding new components to $s_{i-1}$. (This means freezing all used components 
of any $s_k$ once $T_k$ is found.)
By restricting $S$ to self-delimiting prefix codes like those  generated by
the Optimal Ordered Problem Solver (OOPS) \cite{Schmidhuber:04oops}, 
one can now profit from a sometimes 
particularly efficient type of {\sc Correctness Demonstration},
ensuring that  differences between $s_i$ and $s_{i-1}$ cannot affect solutions to $T_{< i}$ under
certain conditions.
More precisely, to obtain $s_i$, half the search time is spent on trying to process $T_i$ first by $s_{i-1}$, 
extending or prolonging $s_{i-1}$
only when the ongoing computation requests to add new components 
through special instructions \cite{Schmidhuber:04oops}---then {\sc Correctness Demonstration}
has less to do as the set $T_{< i}$ is guaranteed to remain solvable, by induction. The other half of the time is spent 
on processing $T_i$ by a new sub-program with new components $s'_i$, a part of $s_i$ but {\em not} of $s_{i-1}$, 
where  $s'_i$ may read $s_{i-1}$ or invoke parts of $s_{i-1}$ as sub-programs to solve $T_{\leq i}$ --- only then {\sc Correctness Demonstration} has to test $s_i$ not only on $T_i$ but also on  $T_{< i}$ (see
\cite{Schmidhuber:04oops} for details).

A simple but not very general way of doing something similar is to 
interleave {\sc Task Invention}, {\sc Solver Modification}, {\sc Correctness Demonstration} as follows:
restrict all $p \in \cal P$ such that they must define $I_i:=i$ as the unique task identifier $I_i$ for $T_i$ (see Section \ref{decisionmaking});
restrict all $s \in \cal S$ such that  the input of $I_i=i$ automatically  invokes sub-program $s'_i$, a part of $s_i$ but {\em not} of $s_{i-1}$ 
(although  $s'_i$ may read $s_{i-1}$ or invoke parts of $s_{i-1}$ as sub-programs to solve $T_i$).
Restrict $J_i$ to a subset of acceptable computational outcomes (Section \ref{decisionmaking}).
Run $s_i$ until it halts and has computed a {\em novel} output acceptable by $J_i$ that is 
different from all outputs computed
by the (halting) solutions to $T_{< i}$;
this novel output becomes $T_i$'s goal.
By induction over $i$,
since  all previously used components of $s_{i-1}$ remain unmodified, the set $T_{< i}$ 
is guaranteed to remain solvable, no matter $s'_i$.  That is, {\sc Correctness Demonstration} on previous tasks
becomes trivial. However, in this simple setup 
there is no immediate generalization across tasks like in OOPS \cite{Schmidhuber:04oops} and the previous paragraph:
the trivial task identifier
$i$ will always first invoke some $s'_i$ different from all $s'_k (k\neq i)$, instead of allowing for solving a new task 
solely by previously found code.

\section{Implementations of {\sc PowerPlay}}
\label{implement}

{\sc PowerPlay} is a general framework that allows for plugging in many differents search and learning algorithms.
The present section will discuss some of them.

\subsection{Implementation Based on Optimal Ordered Problem Solver OOPS}
\label{oops}

\begin{algorithm}{\bf Alg. \ref{dooops}: Implementing {\sc PowerPlay} with Procedure OOPS \cite{Schmidhuber:04oops}}
\label{dooops}
\begin{algorithmic}
\STATE (see text for details) - initialize $s_0$ and 
$u$ (internal dynamic storage for $s \in \cal S$) and $U$ (internal dynamic storage for $p \in \cal P$),
where each possible $p$ is a sequence  of subprograms  $p',p'',p'''$.
\FOR {$i := 1, 2, \ldots$} 
\STATE set variable time limit $t_{lim}:=1$; 
\STATE let the variable set $H$ be empty;
\STATE set Boolean variable DONE $:=$ FALSE
\REPEAT 
\IF {$H$ is empty} 
\STATE set $t_{lim} := 2 t_{lim}$; $H := \{p \in \cal P$  $: P(p)t_{lim}  \geq 1\}$
\ELSE
\STATE choose and remove some   $p$ from $H$ 
\WHILE {not DONE and less than $P(p)t_{lim}$ time was spent on $p$}
\STATE execute the next time step of the following computation:
\STATE 1. Let $p'$ compute some task $T \in \cal T$ and halt.
\STATE 2. Let $p''$ compute $q \in \cal S$ by modifying a copy of $s_{i-1}$, and halt.
\STATE 3. Let $p'''$ try to show that  $q$ but not $s_{i-1}$ can solve $T_1, T_2, \ldots, T_{i-1}, T$. 
\STATE \ \ \ \ \ If $p'''$ was successful set DONE $:=$ TRUE. 
\ENDWHILE
\STATE Undo all modifications of $u$ and $U$ due to $p$. This does not cost more time than executing $p$ in the while loop above \cite{Schmidhuber:04oops}.
\ENDIF
\UNTIL {DONE}
\STATE set $p_i := p$; $T_i := T$; $s_i := q$;
\STATE add a unique encoding of the 5-tuple $(i, p_i, s_i, T_i, Trace_i)$ to read-only storage readable by 
programs to be tested in the future.
\ENDFOR
\end{algorithmic}
\end{algorithm}

The $i$-th problem is to find a program $p_i \in \cal P$ that creates  $s_i$ and  $T_i$ 
and demonstrates that  $s_i$ but not $s_{i-1}$ can solve $T_1, T_2, \ldots, T_i$.
This yields a perfectly ordered problem sequence for a variant of
the {\em Optimal Ordered Problem Solver}
 OOPS \cite{Schmidhuber:04oops} (Algorithm \ref{dooops}).
 
While  a candidate program $p \in \cal P$ is executed,
at any given discrete time step $t=1,2,... $,
its internal state or dynamical storage $U$ at time $t$ is denoted $U(t) \in B^*$
(not to be confused with the solver's internal state $u(t)$ of Section \ref{decisionmaking}).
Its initial default value is $U(0)$. E.g., 
$U(t)$ could encode the current contents of the internal tape of a TM (to be modified by $p$),
or of certain cells in the dynamic storage area of a PC.
 
Once $p_i$ is found, $p_i, s_i, T_i, Trace_i$ (if applicable; see Section \ref{decisionmaking}) 
will be saved in unmodifiable read-only storage,
possibly together with other data observed during the search so far. This
may greatly facilitate the search for  $p_k, k>i$, since
$p_k$ may contain instructions for addressing and reading $p_j, s_j, T_j, Trace_j (j=1, \ldots,k-1)$  and for copying the read code
into {\em modifiable} storage $U$, where $p_k$ may further edit the code, and execute the result, which may be a useful subprogram \cite{Schmidhuber:04oops}.

Define a probability distribution $P(p)$ on $\cal P$  to represent the searcher's 
initial bias (more likely programs $p$ will be tested earlier \cite{Levin:73}). $P$ could be based on program length, e.g., 
$P(p)=2^{-L(p)}$, or on a 
probabilistic syntax diagram \cite{Schmidhuber:04oops,Schmidhuber:04oopscode}.
See Algorithm \ref{dooops}.

OOPS keeps doubling the time limit until there is sufficient runtime for a sufficiently likely program 
to compute a novel, previously unsolvable task, plus its solver, which provably does not forget previous solutions.
OOPS  allocates time to programs according to 
an asymptotically optimal universal search method \cite{Levin:73}  for
problems with easily verifiable solutions, that is, solutions
whose validity can be quickly tested.
Given some problem class,
if some unknown optimal program $p$ requires $f(k)$ steps to solve a
problem instance of size $k$ and demonstrate the correctness of the result,
then this search method
will need at most $O(f(k) / P(p)) = O(f(k))$ steps---the constant 
factor  $1/P(p)$ may be large but does not depend on $k$.
Since OOPS may re-use previously generated solutions
and solution-computing programs, however, it may be possible to greatly 
reduce the constant factor  associated with plain universal search
\cite{Schmidhuber:04oops}.

The big difference to previous implementations of OOPS is that 
{\sc PowerPlay} has the additional freedom to define its own tasks. 
As always, every new task added to the repertoire is essentially defined by the time required 
to invent it, to solve it, and to demonstrate that no previously learned skills got lost.

\subsubsection{Building on Existing OOPS Source Code }
\label{forth}

Existing OOPS source code
\cite{Schmidhuber:04oopscode} uses a FORTH-like universal programming language
to define $\cal P$. It already contains a framework for testing new code on
previously solved tasks, and for efficiently undoing all $U$-modifications of each tested program. 
The source code will require a few changes to 
implement  the additional task search described above.

\subsubsection{Alternative Problem Solvers Based on Recurrent Neural Networks}
\label{rnn}

Recurrent NNs (RNNs, e.g., \cite{Werbos:88gasmarket,WilliamsZipser:92,RobinsonFallside:87tr,Schmidhuber:92ncn3,Hochreiter:97lstm})
are general computers that allow for both sequential
and parallel computations, unlike the strictly sequential FORTH-like language of Section \ref{forth}. 
They can compute any function computable by a standard PC \cite{Schmidhuber:90thesis}.
The original report \cite{Schmidhuber:11powerplay}
used a fully connected RNN called RNN1 to define $\cal S$, where
$w^{lk}$ is the real-valued {\em weight} on the directed connection
between the $l$-th and $k$-th neuron. To program RNN1 means to set the weight matrix $s = \langle w^{lk} \rangle$.
Given enough neurons with appropriate activation functions and an appropriate $\langle w^{lk} \rangle$,
 Algorithm \ref{dooops} can be used to train $s$. $\cal P$ may itself be the set of
weight matrices of a separate RNN called RNN2, computing tasks for RNN1, 
and modifications of RNN1, using techniques for network-modifying networks
as described in previous work \cite{Schmidhuber:92ncfastweights,Schmidhuber:93selfreficann,Schmidhuber:93ratioicann}.

In first experiments  \cite{rupesh2012icdl,powerplay2012experiments},
a particularly suited NN called a self-delimiting NN or SLIM NN \cite{Schmidhuber:12slimnn} is used.
During program execution or activation spreading in the SLIM NN, lists are used to trace only those neurons and connections used at least once.
This also allows for efficient resets of large NNs which  may use only a small fraction of their weights per task.
Unlike standard RNNs, 
SLIM NNs are easily combined with techniques of asymptotically optimal program search \cite{Levin:73,Schmidhuber:97bias,Schmidhuber:03nips,Schmidhuber:04oops} 
 (Section \ref{oops}).
To address overfitting, instead of depending on  pre-wired regularizers and hyper-parameters \cite{bishop:2006},
SLIM NNs  can in principle learn to select by themselves their own runtime and their own numbers of free parameters,
 becoming fast and  {\em slim} when necessary.
Efficient SLIM NN learning algorithms (LAs) track which weights are used for which tasks  (Section \ref{components}),
to greatly speed up performance evaluations in response to limited weight changes.
LAs may penalize the task-specific  total length of connections used by 
SLIM NNs implemented on the 3-dimensional brain-like multi-processor hardware to expected in the future.
This  encourages SLIM NNs to solve many subtasks by subsets of neurons that are physically close \cite{Schmidhuber:12slimnn}.

\subsection{Adapting the Probability Distribution on Programs}
A straightforward extension of the above works as follows: Whenever a new
$p_i$ is found, $P$ is updated to make either only $p_i$ or
all $p_1,p_2, \ldots, p_i$ more likely. Simple ways of doing this 
are described in previous work \cite{Schmidhuber:97bias}.
This may be justified to the extent that future successful programs 
turn out to be similar to previous ones.

\subsection{Implementation Based on Stochastic or Evolutionary Search}
\label{evolution}

A possibly simpler but less general approach is to use
an evolutionary algorithm
to produce an $s$-modifying and task-generating program $p$ as requested by {\sc PowerPlay},
according to Algorithm \ref{doevo}, which refers to the recurrent net problem solver of Section \ref{rnn}.

\begin{algorithm}{\bf Alg. \ref{doevo}: {\sc PowerPlay} for RNNs Using Stochastic or Evolutionary Search}
\label{doevo}
\begin{algorithmic}
\STATE Randomly initialize RNN1's variable weight matrix $\langle w^{lk} \rangle$ and use the result as $s_0$ (see Section \ref{rnn})
\FOR {$i := 1, 2, \ldots$} 
\STATE set Boolean variable DONE$=$FALSE
\REPEAT 
\STATE use a black box optimization algorithm BBOA (many are possible
 \cite{Rechenberg:71,Gomez:08jmlr,wierstraCEC08,sehnke2009parameter}) with adaptive parameter vector $\theta$  to create some $T \in \cal T$ (to define the task input to RNN1; see Section
\ref{taskinvention}) and a modification of $s_{i-1}$, the current $\langle w^{lk} \rangle$ of RNN1, thus obtaining a new candidate $q \in \cal S$
\IF {$q$ but not $s_{i-1}$ can solve $T$ and all $T_k (k<i)$  (see Sections \ref{correctness}, \ref{components})} 
\STATE set DONE$=$TRUE
\ENDIF
\UNTIL {DONE}
\STATE set $s_i := q$; $\langle w^{lk} \rangle := q$; $T_i := T$; (also store $Trace_i$ if applicable, see Section  \ref{decisionmaking}). Use the information stored so far to adapt the parameters $\theta$ of the BBOA, e.g., by gradient-based search  \cite{wierstraCEC08,sehnke2009parameter},  or according to the principles of evolutionary computation \cite{Rechenberg:71,Gomez:08jmlr,wierstraCEC08}.
\ENDFOR
\end{algorithmic}
\end{algorithm}

\section{Outgrowing Trivial Tasks - Compressing Previous Solutions}
\label{compressing}

What prevents {\sc PowerPlay} from inventing trivial  tasks forever by extreme modularization, simply allocating a previously unused solver part to each new task, which thus becomes rather quickly verifiable, as its solution does not affect solutions to previous tasks (Section  \ref{prefix})?
On realistic but general architectures such as PCs and RNNs,
at least once the upper storage size limit of $s$ is reached, {\sc PowerPlay} will 
start ``compressing'' previous solutions, making $s$ {\em generalize} in the sense 
that the {\em same} relatively short piece of code (some part of $s$) helps to solve {\em different}
tasks. 

With many computational architectures,
this type of compression 
will start much earlier though, because new tasks solvable
by partial reuse of earlier discovered code will often be easier to find than new tasks 
solvable by previously unused parts of $s$. This also holds for
growing architectures with potentially unlimited storage space.

Compare also {\sc PowerPlay} Variant II of Section  \ref{costsection}  
whose tasks may explicitly require improving the {\em average}
 time and space complexity of previous solutions by some minimal value.

In general, however, over time the system will find it more and more difficult to invent
novel tasks without forgetting previous solutions, a bit like humans find it harder
and harder to learn truly novel behaviors once they are leaving behind the
initial rapid exploration phase typical for babies.
Experiments with various problem solver architectures (e.g., \cite{rupesh2012icdl,powerplay2012experiments})
 are needed to analyze such effects in detail.

\section{Adding External Tasks}
\label{external}

The growing repertoire of the problem solver  may facilitate learning of solutions to externally posed tasks.
For example, one may modify {\sc PowerPlay} such that for certain $i$, $T_i$ is defined externally,
instead  of being invented by the system itself. In general, the resulting $s_i$ will contain an
externally  inserted bias in
form of code that will make some future self-generated tasks easier to find than others.
It should be possible to push the system in a human-understandable or otherwise useful
direction by regularly inserting appropriate external goals. See Algorithm \ref{cost}.

Another way of exploiting the growing repertoire is to simply copy $s_i$ for some $i$
and use it as a starting point for a search for a solution to an externally posed task $T$, {\em without}
insisting that the modified $s_i$ also can solve $T_1, T_2, \ldots, T_i$. This may be much faster
than trying to solve $T$ from scratch, to the extent the solutions to self-generated
tasks reflect general knowledge (code) re-usable for $T$.

In general, however, it will be possible to design external tasks 
whose solutions do {\em not} profit from those of
self-generated tasks---the latter even may turn out to slow down the search.

On the other hand, in the real world the benefits of curious exploration
seem obvious. One should analyze theoretically and experimentally under which conditions 
the creation of self-generated tasks can accelerate the solution to externally generated
tasks---see \cite{Schmidhuber:91singaporecur,Storck:95,Schmidhuber:99cec,Schmidhuber:02predictable,luciw:icdl2011,cuccu2011,Schaul2011cogp,sunyi2011agi}
for previous simple experimental studies in this vein.

\subsection{Self-Reference Through Novel Task Search as an External Task}
\label{self}

 {\sc PowerPlay}'s $i$-th goal is to find a $p_i \in \cal P$ that creates $T_i$ and $s_i$ 
(a modification of $s_{i-1}$)    
and shows that  $s_i$ but not $s_{i-1}$ can solve $T_{\leq i}$. 
As $s$ itself is becoming a more and more general problem solver,  
$s$ may help in many ways to achieve such goals in self-referential fashion.
For example, the old solver $s_{i-1}$ may be able to
read a unique formal description (provided by $p_i$) of {\sc PowerPlay}'s $i$-th goal, viewing it as an external task,
and produce an output unambiguously describing
a candidate for ($T_i,s_i$). 
If $s$ has a theorem prover component (Section \ref{proofsearch}),
$s_{i-1}$ might even output a full proof of ($T_i,s_i$)'s validity; alternatively $p_i$ 
could just use the possibly suboptimal
suggestions of $s_{i-1}$ to  narrow down and speed up the search,
one of the reasons why Section \ref{framework} already mentioned that
programs $p \in \cal P$
should contain instructions for reading  (and running) the code of the present problem solver.

\section{Softening Task Acceptance Criteria of {\sc PowerPlay}}
\label{softening}

The {\sc PowerPlay} variants above insist that $s$ may not solve new tasks at the expense of forgetting to solve any previously solved task
within its previously established time and space bounds. 
For example, let us consider the sequential decision-making tasks from Section \ref{decisionmaking}.
Suppose the problem solver can already solve task $T_k=(I_k, J_k, t_k, n_k)\in \cal I \times \cal J \times \mathbb{N} \times \mathbb{N}$.
A very similar but admissible new task  $T_i=(I_k, J_k, t_i, n_k), (i>k),$ would be to solve $T_k$ substantially
faster: $t_i<t_k - \epsilon$,
as long as $T_i$ is not already solvable by $s_{i-1}$,
and no solution to some $T_l(l<i)$ is forgotten in the process. 

Here I discuss variants of {\sc PowerPlay} that soften the acceptance criteria for new tasks in various ways,
 for example, by allowing some of the computations of solutions to previous non-external (Section \ref{external}) tasks to slow down by a certain amount of time, provided the {\em sum} of their runtimes  does not decrease.
This also permits the system to invent new previously unsolved tasks at the expense of slightly increasing time bounds 
for certain already solved non-external tasks, but without decreasing the average performance on the latter.
Of course, {\sc PowerPlay} has to be modified accordingly, updating average runtime bounds when necessary.

Alternatively, one may allow for trading off space and time constraints in reasonable ways, e.g., in
the style of asymptotically optimal {\em Universal Search} \cite{Levin:73}, which essentially  trades one bit of 
additional space complexity for a runtime speedup factor of 2. 

\subsection{{\sc PowerPlay} Variant II: Explicitly Penalizing Time and Space Complexity}
\label{costsection}

Let us remove time and space bounds from the task definitions of Section \ref{decisionmaking},  
since the modified cost-based {\sc PowerPlay} framework below (Algorithm \ref{cost}) will handle 
computational costs (such as time and space complexity of solutions) more directly.
In the present section, $T_i$ encodes a tuple $(I_i, J_i)\in \cal I \times \cal J$ 
with interpretation:
$s_i$ must  first read $I_i$ 
and then interact with an environment through a sequence of perceptions and actions,
to achieve some computable goal defined by $J_i$ within a certain maximal time interval $t_{max}$ (a positive constant).
Let $t'_s(T)$  be  $t_{max}$ if $s$ cannot solve task $T$, otherwise it is the time needed to solve $T$ by $s$.
Let $l'_s(T)$  be the positive constant $l_{max}$ if $s$ cannot solve $T$, 
otherwise it is the number of components of $s$ needed to solve task $T$ by $s$.
The non-negative real-valued reward $r(T)$ for solving $T$  is a positive constant $r_{new}$ for self-defined 
previously unsolvable $T$,
or user-defined if  $T$ is an external task solved by $s$ (Section \ref{external}). 
The real-valued cost $Cost(s, TSET)$ of solving all tasks in a  task set $TSET$  through $s$
is a real-valued function of:  all $l'_s(T)$, $t'_s(T)$ (for all $T \in TSET$), $L(s)$,
 and $\sum_{T \in TSET} r(T)$.
For example, the cost function  $Cost(s, TSET)=  L(s) +  \alpha \sum_{T \in TSET} [ t'_s(T)-r(T)]$
encourages compact and fast solvers solving many different tasks with the same components of $s$,
where the real-valued positive parameter $\alpha$ weighs space costs against time costs, and
$r_{new}$ should exceed $t_{max}$ to encourage solutions of novel self-generated tasks,
whose cost contributions should be below zero
(alternative cost definitions could also take into account energy consumption etc.)

Let us keep an analogue of the remaining notation of Section \ref{decisionmaking},
such as $u_i(t),x_i(t), r_i(t),y_i(t),Trace_i,\\ J_i (Trace_i)$. As always, if the environment 
is unknown and possibly changing over time, to test performance 
of a new solver $s$ on a previous task $T_k$, only $Trace_k$ is necessary---see Section \ref{decisionmaking}.
As always, let $T_{\leq i}$ denote the set containing all tasks $T_1, \ldots, T_i$ (note that if $T_i$=$T_k$ for 
some $k<i$ then it will appear only once in $T_{\leq i}$), and
let  $\epsilon>0$ again define what's acceptable progress:

\begin{algorithm}{\bf Alg. \ref{cost}:   {\sc PowerPlay} Framework (Variant II) Explicitly Handling Costs of Solving Tasks}
\label{cost}
\begin{algorithmic}
\STATE Initialize $s_0$ in some way
\FOR {$i := 1, 2, \ldots$} 
\STATE Create new global variables $T_i \in \cal T$,  $s_i \in \cal S$,  $p_i  \in \cal P$, $c_i, c^*_i \in \mathbb{R}$ (to be fixed by the end of {\bf repeat})
\REPEAT 
\STATE Let a search algorithm (Section \ref{oops}) set $p_i$, a new candidate program. Give $p_i$ limited time to do:
\STATE  {\bf *} {\sc Task Invention}: Unless the user specifies  $T_i$ (Section \ref{external}), let $p_i$ set  $T_i$.
\STATE   {\bf *}  {\sc Solver Modification}: Let $p_i$ set $s_i$  by computing a modification of $s_{i-1}$ (Section \ref{policychange}).
\STATE   {\bf *}  {\sc Correctness Demonstration}: 
Let $p_i$  compute $c_i := Cost(s_i, T_{\leq i})$ and $c^*_i := Cost(s_{i-1}, T_{\leq i})$
\UNTIL {$c^*_i - c_i > \epsilon$ (minimal savings of costs such as time/space/etc on all tasks so far)}
\STATE Freeze/store forever $p_i, T_i, s_i, c_i, c^*_i$ 
\ENDFOR
\end{algorithmic}
\end{algorithm}

By Algorithm \ref{cost}, $s_i$ may forget certain abilities of $s_{i-1}$, provided that the overall performance as measured by $Cost(s_i, T_{\leq i})$
has improved, either because a new task became solvable, or previous tasks became solvable more efficiently.

Following Section \ref{correctness},  {\sc Correctness Demonstration} can often be facilitated,
for example, 
by tracking which components of $s_i$ are used for solving which tasks  (Section \ref{components}).

To further refine this approach, consider that in phase $i$, the list
$L^k_i$ (defined in Section \ref{components}) contains all previously learned tasks whose solutions depend on  $s^k$.
This  can be used to determine the current {\em value} $Val(s^k_i)$ of some component $s^k$ of $s$:
$Val(s^k_i)= - \sum_{T \in L^k_i} Cost(s_i, T_{\leq i})$. It is a simple exercise to invent {\sc PowerPlay} variants that do not
forget valuable components as easily as less valuable ones.

The  implementations of
Sections \ref{oops} and \ref{evolution} are easily adapted to the cost-based {\sc PowerPlay} framework.
Compare separate papers \cite{rupesh2012icdl,powerplay2012experiments}.

\subsection{Probabilistic {\sc PowerPlay} Variants}
\label{probabilistic}

Section \ref{decisionmaking} pointed out that in partially observable and/or
non-stationary unknown environments 
{\sc Correctness Demonstration} must use $Trace_k$ to check whether a new $s_i$ still knows
how to solve an earlier task $T_k (k<i)$.
A less strict variant of {\sc PowerPlay}, however, will simply make certain
assumptions about the probabilistic nature of the environment and the repeatability of  trials,
assuming that a limited fixed number of interactions with the real world are sufficient to
estimate the costs $c^*_i, c_i$ in Algorithm \ref{cost}.

Another probabilistic way of softening {\sc PowerPlay} is to add new tasks without proof that $s$ won't forget solutions to previous tasks, provided  {\sc Correctness Demonstration} can at least show that the probability of forgetting any previous solution
 is below some real-valued positive constant threshold.

\section{First Illustrative Experiments}
\label{experiments}

First experiments are reported in separate papers \cite{rupesh2012icdl,powerplay2012experiments}
(some experiments were also briefly mentioned in the original report \cite{Schmidhuber:11powerplay}).
Standard NNs as well as
SLIM RNNs  \cite{Schmidhuber:12slimnn}
are used as computational problem solving 
  architectures.  The weights of SLIM RNNs can encode essentially arbitrary computable tasks as well 
as arbitrary, self-delimiting, halting or
  non-halting programs solving those tasks. These programs may affect both environment (through effectors) and
  internal states encoding abstractions of event sequences.  In open-ended
  fashion, the {\sc PowerPlay}-driven NNs learn to become  increasingly
  general solvers of self-invented problems, continually adding new problem solving procedures to
  the growing repertoire, sometimes compressing/speeding up previous skills,
sometimes preferring to invent new tasks and corresponding skills. The NNs exhibit interesting developmental stages.
It is also shown how a {\sc PowerPlay}-driven SLIM NN  automatically self-modularizes \cite{powerplay2012experiments}, 
frequently re-using code for previously invented skills,
always trying to invent novel tasks that can be quickly validated because they do not require too 
many weight changes affecting too many previous tasks.

\section{Previous Relevant Work}
\label{previous}

Here I discuss related research, in particular, why the present work is of interest despite
the recent advent of theoretically optimal universal problem solvers (Section \ref{optimal}),
and how it can be viewed as a greedy but feasible and sound implementation
of the formal theory of creativity (Section \ref{creativity}).

\subsection{Existing Theoretically Optimal Universal Problem Solvers}
\label{optimal}

The new millennium brought universal problem solvers
that are theoretically optimal in a certain sense.
The fully self-referential \cite{Goedel:31}
{\em \gm } \cite{Schmidhuber:05gmai,Schmidhuber:09gm}  may
interact with some initially unknown, partially observable environment
to maximize future expected utility or reward by solving arbitrary
user-defined computational tasks.
Its initial algorithm is not hardwired;
it can completely rewrite itself without essential limits
apart from the limits of computability,
but only if a proof searcher embedded within the initial algorithm 
can first prove that the rewrite is useful, 
according to the formalized utility function taking into account the
limited computational resources.
Self-rewrites due to this approach can be shown to be
{\em globally optimal}, relative to G\"{o}del's 
well-known fundamental restrictions of provability \cite{Goedel:31}.
To make sure the \gm is at least {\em asymptotically} optimal even before the first self-rewrite,
one may initialize it by
Hutter's non-self-referential but 
{\em asymptotically fastest algorithm for all well-defined problems}
\hs  \cite{Hutter:01fast+}, which 
uses a hardwired brute force proof searcher  and
ignores the costs of proof search.
Assuming discrete input/output domains $X/Y \subset B^*$, a formal problem
specification $f: X \rightarrow Y$
(say, a functional description of how integers are decomposed
into their prime factors),
and a particular $x \in X$ (say,
an integer to be factorized), \hs
orders all proofs of an appropriate axiomatic system
by size to find programs $q$ that
for all $z \in X$ provably compute $f(z)$
within time bound $t_q(z)$. Simultaneously it
spends most of its time on executing the $q$ with the
best currently proven time bound $t_q(x)$.
 \hs is as fast as the {\em fastest} algorithm that provably
computes $f(z)$ for all $z \in X$, save for a constant factor
smaller than $1 + \epsilon$ (arbitrarily small real-valued $\epsilon > 0$)
and an $f$-specific but $x$-independent
additive constant  \cite{Hutter:01fast+}.
Given some problem, the \gm may decide to replace \hs by a faster method  suffering less from 
large constant overhead,
but even if it doesn't, its performance won't be less than asymptotically optimal.

Why doesn't everybody use such universal problem solvers for all computational
real-world  problems? Because most real-world problems
are so small that the ominous constant slowdowns (potentially relevant at least before
the first self-rewrite) may be large enough to prevent  the universal methods
from being  feasible.

{\sc PowerPlay}, on the other hand, is designed to incrementally build
a  {\em practical} more and more general problem solver that can solve 
numerous tasks quickly, not in the asymptotic sense,
but by exploiting to the max its given particular search algorithm 
and  computational 
architecture, with all its space and time limitations, including those 
reflected by constants  ignored by the asymptotic optimality notation.

As mentioned in Section \ref{external}, however, one must now analyze under which conditions 
{\sc PowerPlay}'s
self-generated tasks can accelerate the solution to externally generated
tasks (compare previous experimental studies of this type \cite{Schmidhuber:91singaporecur,Storck:95,Schmidhuber:99cec,Schmidhuber:02predictable}).

\subsection{Connection to Traditional Active Learning}
\label{active}

Traditional active learning methods \cite{Fedorov:72} such as AdaBoost \cite{Freund1997119} 
have a totally different set-up and purpose: there the user  
provides a set of samples to be learned, then each new classifier in a  
series of classifiers focuses on samples badly classified by previous  
classifiers. Open-ended {\sc PowerPlay}, however, 
(1) considers arbitrary computational problems (not necessarily classification tasks);
(2) can self-invent all computational tasks.
There is no  need for a
pre-defined global set of tasks that each new solver tries to solve  
better, instead the task set continually grows based on which task is  
easy to invent and validate, given what is already known.

\subsection{Greedy Implementation of Aspects of the Formal Theory of Creativity}
\label{creativity}

The Formal Theory of Creativity \cite{Schmidhuber:06cs,Schmidhuber:10ieeetamd}
considers agents living in initially unknown environments. At any given time, 
such an agent uses a reinforcement learning (RL) method \cite{Kaelbling:96} to 
maximize not only expected future external reward for achieving certain goals,
but also {\em intrinsic} reward for improving an internal model of the environmental responses
to its actions,
learning to better predict or compress the growing history of observations
influenced by its behavior,
thus achieving {\em wow-effects},
actively learning skills to influence the input stream such that it contains previously
unknown but learnable algorithmic regularities. 
I have argued that the theory explains essential aspects of intelligence 
including  selective attention, curiosity, creativity, 
science, art, music, humor, e.g.,
 \cite{Schmidhuber:06cs,Schmidhuber:10ieeetamd}.
Compare recent related work, e.g., 
  \cite{Barto:12intrinsic,Dayan:12intrinsic,Oudeyer:12intrinsic,Nemzhov:12intrinsic}.

Like {\sc PowerPlay}, such a creative agent  produces a sequence of self-generated
tasks and their solutions, each task still unsolvable before 
learning, yet becoming solvable after learning. 
The costs of learning as well as the learning progress are measured, 
and enter the reward function. Thus, in the absence of external reward for reaching
user-defined goals,
 at any given time the agent is motivated to 
invent a series of additional tasks that maximize future expected learning progress.

For example, by restricting its input stream to self-generated pairs $(I, O) \in \cal I \times \cal O$ like
in Section \ref{patternrecognition}, and limiting it to predict only $O$, given $I$
(instead of also trying to predict future $(I, O)$ pairs from previous ones, which the general agent would do),
there will be a motivation to actively generate a sequence of $(I, O)$ pairs such that the $O$ are first
subjectively unpredictable from their $I$ but then become predictable with little effort,
given the limitations of whatever learning algorithm is used.

Here some cons and pros of {\sc PowerPlay} are listed in light of the above.
Its drawbacks include:
\begin{enumerate}
\item
Instead of maximizing future expected reward, {\sc PowerPlay}
is greedy, always trying to find the simplest (easiest to find and validate) task to add to the repertoire,
or the simplest way of improving the efficiency or compressibility of previous solutions,
instead of looking further ahead, as a universal RL method \cite{Schmidhuber:06cs,Schmidhuber:10ieeetamd} would do.
That is, {\sc PowerPlay} may potentially sacrifice large long-term gains for small short-term gains: 
the discovery of many easily solvable tasks may at least temporarily prevent it from learning to solve hard tasks. 

On general computational architectures such as RNNs (Section \ref{rnn}), however, 
{\sc PowerPlay} is expected to soon run out of easy tasks that are not yet solvable,
due to the architecture's limited capacity and its unavoidable generalization effects 
(many never-tried tasks will become solvable by solutions to the few 
explicitly tested $T_i$). Compare Section \ref{compressing}.
\item
The general creative agent above \cite{Schmidhuber:06cs,Schmidhuber:10ieeetamd} is motivated to improve performance on the  
entire history of previous still unsolved tasks, while {\sc PowerPlay} may discard much of this history, keeping only a selective list of previously solved tasks. However, as the system is interacting with its environment, one could store the entire continually growing
history, and make sure that $\cal T$ always allows for defining the task of better compressing the history so far.
\item
{\sc PowerPlay} as in Section \ref{framework} has a binary criterion for  adding knowledge (was the new task solvable without forgetting old solutions?), while the general agent \cite{Schmidhuber:06cs,Schmidhuber:10ieeetamd} uses a more informative information-theoretic measure. The cost-based {\sc PowerPlay}  framework (Alg. \ref{cost}) of Section \ref{softening}, however, offers similar, more flexible options,
rewarding compression or speedup of solutions to previously solved tasks.
\end{enumerate}

\noindent
On the other hand, drawbacks of previous implementations of formal creativity  theory  include:
\begin{enumerate}
\item
Some previous approximative implementations  \cite{Schmidhuber:91singaporecur,Storck:95} 
used traditional RL methods \cite{Kaelbling:96} with theoretically unlimited look-ahead, 
but those are not guaranteed
to work well in partially observable and/or non-stationary environments 
where the reward function changes over time, and won't necessarily
generate an optimal sequence of future tasks or experiments. 
\item
Theoretically optimal implementations
\cite{Schmidhuber:06cs,Schmidhuber:10ieeetamd} are currently still impractical,
for reasons similar to those discussed in Section \ref{optimal}.
\end{enumerate}

Hence {\sc PowerPlay} may be viewed as a  greedy
but  feasible implementation of certain basic principles of creativity \cite{Schmidhuber:06cs,Schmidhuber:10ieeetamd}.
{\sc PowerPlay}-based systems are
continually motivated to invent new tasks solvable by formerly unknown procedures, 
or to compress or speed up problem solving procedures discovered earlier.
Unlike previous implementations,
{\sc PowerPlay} extracts from the lifelong experience history a sequence of clearly identified and separated tasks
with explicitly recorded solutions. By design it cannot suffer from online learning problems 
affecting its solver's
performance on previously solved problems.

\subsection{Beyond Algorithmic Zero-Sum Games \cite{Schmidhuber:99cec,Schmidhuber:02predictable} (1997-2002)}
\label{zerosum}

This guaranteed
robustness against forgetting previous skills also represents a difference to the most closely related
 previous work \cite{Schmidhuber:99cec,Schmidhuber:02predictable}.
There, to address the computational costs of learning, and the costs of measuring learning progress,
computationally powerful encoders and problem solvers 
 \cite{Schmidhuber:97interesting,Schmidhuber:02predictable}  (1997-2002) are implemented 
as two very general, co-evolving, symmetric, opposing modules called the {\em right brain}
and the {\em left brain}. Both are able to construct self-modifying
probabilistic programs written in a universal programming language.
An internal storage for temporary computational results
of the programs is viewed as part of the changing environment.
Each module can suggest experiments 
in the form of probabilistic algorithms to be executed,
and make predictions about their effects, 
{\em betting intrinsic reward}  on their outcomes.
The opposing module may accept such a bet in a zero-sum game by making a 
contrary prediction, or reject it. In case of acceptance,
the winner is determined by executing the experiment and 
checking its outcome;
the intrinsic reward eventually gets transferred from the surprised loser to the 
confirmed winner.  
Both modules try to maximize reward using
a rather general RL algorithm (the so-called success-story algorithm SSA  \cite{Schmidhuber:97bias}) 
designed for  complex stochastic policies (alternative
RL algorithms could be plugged in as well).
Thus both modules are motivated to discover {\em novel} algorithmic
patterns/compressibility ($=$ surprising {\em wow-effects}), where the subjective baseline
for novelty is given by what the opponent already knows about
the (external or internal) world's repetitive patterns.
Since the execution of any computational or physical action
costs something (as it will reduce the cumulative reward per time ratio), both modules
are motivated to focus on those parts of the dynamic
world that currently make surprises and learning progress {\em easy}, to minimize
the costs of identifying promising experiments and executing them.
The system learns a  partly hierarchical structure of more and more complex 
skills or programs necessary to solve the growing sequence of self-generated tasks,
reusing previously acquired simpler skills where this is beneficial.
Experimental studies exhibit
several sequential stages of
emergent developmental sequences, with and without external
reward \cite{Schmidhuber:99cec,Schmidhuber:02predictable}.

However, the previous system \cite{Schmidhuber:99cec,Schmidhuber:02predictable} 
did not have a built-in guarantee that it cannot forget previously learned skills,
while {\sc PowerPlay} as in Section \ref{framework} does (and the time and space complexity-based  variant Alg. \ref{cost} of Section \ref{softening} 
can forget only if this improves the average efficiency of
previous solutions).

To analyze the novel framework's consequences 
in practical settings,  experiments are currently  being conducted with various
problem solver architectures with different generalization properties.
See separate papers \cite{rupesh2012icdl,powerplay2012experiments} and
Section \ref{experiments}.

\subsection{Opposing Forces: Improving Generalization Through Compression, Breaking Generalization Through Novelty}
\label{opposing}
Two opposing forces are at work in  {\sc PowerPlay}. On the one hand, the system continually tries to improve 
previously learned skills, by speeding them up, and by compressing the used parameters of the problem solver, reducing its effective size. 
The compression drive tends to improve generalization performance, according to  the principles of  {\em Occam's Razor}
and \emph{Minimum Description Length} (MDL) and  \emph{Minimum Message Length} (MML)
\cite{Solomonoff:64,Kolmogorov:65,Wallace:68,Wallace:87,Solomonoff:78,Rissanen:78,LiVitanyi:97,Hutter:05book+}.
On the other hand, the system also continually tries to invent new tasks that break the generalization capabilities of the present solver.

{\sc PowerPlay}'s time-minimizing search for new tasks automatically manages the trade-off between these opposing forces. 
Sometimes it is easier (because fewer computational resources are required) to invent and solve a completely new, previously unsolvable problem.
Sometimes  it is easier to compress (or speed up) solutions to previously invented problems.

\subsection{Relation to G\"{o}del's Sequence of Increasingly Powerful Axiomatic Systems}
\label{goedel}

In 1931, Kurt G\"{o}del showed that for each sufficiently powerful ($\omega$-) consistent axiomatic system 
there is a statement that must be true but cannot be proven from the axioms through an algorithmic theorem-proving procedure \cite{Goedel:31}.
This unprovable statement can then be added to the axioms, to obtain a more powerful formal theory in which 
new formerly unprovable theorems become provable, without affecting previously provable theorems.

In a sense,  {\sc PowerPlay} is doing something similar. 
Assume the architecture of the solver is a universal computer \cite{Goedel:31,Turing:36}. Its software $s$ can be viewed as a theorem-proving procedure
implementing certain enumerable axioms and computable inference rules. {\sc PowerPlay} continually tries to modify $s$ such that the  previously proven
theorems remain provable within certain time bounds, {\em and} a new previously unprovable theorem becomes provable.

\section{Words of Caution}
\label{caution}

The behavior of {\sc PowerPlay} is determined by the nature and the limitations of $\cal T$, $\cal S$, $\cal P$, 
and its algorithm for searching $\cal P$. If $\cal T$ includes all computable task descriptions,
and both $\cal S$ and $\cal P$ allow for implementing arbitrary
programs, and the search algorithm is a general method for search in program space (Section \ref{implement}),
then there are few limits  to what {\sc PowerPlay} may do (besides the limits of computability \cite{Goedel:31}).

It may not be advisable to let a general variant of {\sc PowerPlay} loose in an 
uncontrolled situation, e.g., on a multi-computer network on the internet, 
possibly with access to control of physical devices, and the potential to
acquire additional computational and physical resources (Section \ref{decisionmaking})
 through programs executed during {\sc PowerPlay}.
Unlike, say, traditional virus programs, {\sc PowerPlay}-based systems will continually change  in a way 
hard to predict, incessantly inventing and solving novel, self-generated 
tasks, only driven by a desire to increase their general problem-solving capacity,
perhaps a bit like many humans seek to increase their power once their basic needs are satisfied.
This type of artificial curiosity/creativity, however, may conflict with human intentions on occasion.
On the other hand, unchecked curiosity may sometimes also be harmful or fatal 
to the learning system itself (Section \ref{external})---curiosity can kill the cat.

\section{Acknowledgments}
\label{ack}
Thanks to Mark Ring, 
Bas Steunebrink, Faustino Gomez, Sohrob Kazerounian, Hung Ngo, Leo Pape, Giuseppe Cuccu, for useful comments.

\bibliography{bib}
\bibliographystyle{plain}
\end{document}